\documentclass[10pt,twocolumn,letterpaper]{article}

\usepackage[pagenumbers]{cvpr} 

\usepackage{graphicx}
\usepackage{amsmath}
\usepackage{amssymb}
\usepackage{booktabs}
\usepackage{multirow}

\usepackage[pagebackref,breaklinks,colorlinks]{hyperref}

\usepackage[capitalize]{cleveref}
\usepackage{bm}
\crefname{section}{Sec.}{Secs.}
\Crefname{section}{Section}{Sections}
\Crefname{table}{Table}{Tables}
\crefname{table}{Tab.}{Tabs.}

\def\cvprPaperID{5112} 
\def\confName{CVPR}
\def\confYear{2023}

\begin{document}

\title{Zero-Shot Text-to-Parameter Translation for Game Character Auto-Creation}
\author{Rui Zhao\textsuperscript{1},\enspace 
Wei Li\textsuperscript{2},\enspace
Zhipeng Hu\textsuperscript{1},\enspace
Lincheng Li\textsuperscript{1}\footnotemark[1],\enspace
Zhengxia Zou\textsuperscript{3}\footnotemark[1],\enspace
Zhenwei Shi\textsuperscript{3},\enspace
Changjie Fan\textsuperscript{1}\\
\textsuperscript{1}Netease Fuxi AI Lab,\enspace
\textsuperscript{2}Nankai University,\enspace
\textsuperscript{3}Beihang University}

\twocolumn[
{%
\renewcommand\twocolumn[1][]{#1}%
\maketitle
\begin{center}
    \centering
    \includegraphics[width=1.\textwidth]{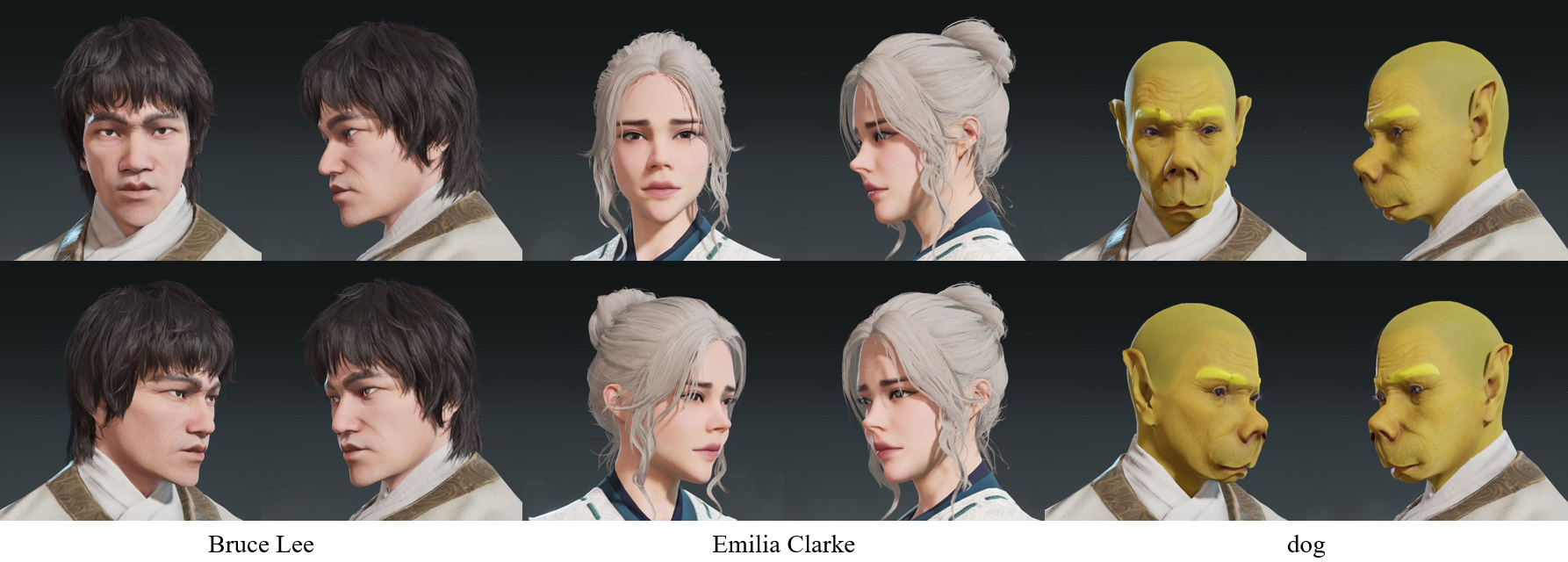}
    \captionof{figure}{Game characters created by the proposed text-to-parameter translation (T2P) given different text prompts. The front view and three side views are shown for each character.}
    \label{fig:teaser}
\end{center}
\vspace{1.6em}
}]

\renewcommand{\thefootnote}{\fnsymbol{footnote}}
\footnotetext[1]{Corresponding Authors.}
\renewcommand{\thefootnote}{\arabic{footnote}}

\begin{abstract}
Recent popular Role-Playing Games (RPGs) saw the great success of character auto-creation systems. The bone-driven face model controlled by continuous parameters (like the position of bones) and discrete parameters (like the hairstyles) makes it possible for users to personalize and customize in-game characters. Previous in-game character auto-creation systems are mostly image-driven, where facial parameters are optimized so that the rendered character looks similar to the reference face photo. This paper proposes a novel text-to-parameter translation method (T2P) to achieve zero-shot text-driven game character auto-creation. With our method, users can create a vivid in-game character with arbitrary text description without using any reference photo or editing hundreds of parameters manually. In our method, taking the power of large-scale pre-trained multi-modal CLIP and neural rendering, T2P searches both continuous facial parameters and discrete facial parameters in a unified framework. Due to the discontinuous parameter representation, previous methods have difficulty in effectively learning discrete facial parameters. T2P, to our best knowledge, is the first method that can handle the optimization of both discrete and continuous parameters. Experimental results show that T2P can generate high-quality and vivid game characters with given text prompts. T2P outperforms other SOTA text-to-3D generation methods on both objective evaluations and subjective evaluations.
\end{abstract}

\section{Introduction}
\label{sec:intro}

Role-Playing Games (RPGs) are praised by gamers for providing immersive experiences. Some of the recent popular RPGs, like Grand Theft Auto Online\footnote[1]{https://www.rockstargames.com/GTAOnline} and Naraka\footnote[2]{http://www.narakathegame.com}, have opened up character customization systems to players. In such systems, in-game characters are bone-driven and controlled by continuous parameters, like the position, rotation, scale of each bone, and discrete parameters, like the hairstyle, beard styles, make-ups, and other facial elements. By manually adjusting these parameters, players can control the appearance of the characters in the game according to their personal preferences, rather than using predefined character templates. 
However, it is cumbersome and time-consuming for users to manually adjust hundreds of parameters - usually taking up to hours to create a character that matches their expectations.

To automatically create in-game characters, the method named Face-to-parameter translation (F2P) was recently proposed to automatically create game characters based on a single input face image~\cite{shi2019face}. F2P and its variants~\cite{shi2020fast, shi2020neural} have been successfully used in recent RPGs like Narake and Justice, 
and virtual meeting platform Yaotai. 
Recent 3D face reconstruction methods~\cite{ booth20173d, peng2017parametric, dou2017end, richardson2017learning, tewari2017mofa, tuan2017regressing, tran2018extreme} can also be adapted to create game characters. However, all the above-mentioned methods require reference face photos for auto-creation. Users may take time to search, download and upload suitable photos for their expected game characters. Compared with images, text prompts are more flexible and time-saving for game character auto-creation. A very recent work AvatarCLIP~\cite{hong2022avatarclip} achieved text-driven avatar auto-creation and animation. It optimizes implicit neural networks to generate characters. However, the created characters are controlled by implicit parameters, which lack explicit physical meanings, thus manually adjusting them needs extra designs. This will be inconvenient for players or game developers to further fine-tune the created game characters as they want. 

To address the above problems, we propose text-to-parameter translation (T2P) to tackle the in-game character auto-creation task based on arbitrary text prompts. T2P takes the power of large-scale pre-trained CLIP to achieve zero-shot text-driven character creation and utilizes neural rendering to make the rendering of in-game characters differentiable to accelerate the parameters optimization. Previous works like F2Ps give up controlling discrete facial parameters due to the problem of discontinuous parameter gradients. To our best knowledge, the proposed T2P is the first method that can handle both continuous and discrete facial parameters optimization in a unified framework to create vivid in-game characters. F2P is also the first text-driven automatic character creation suitable for game environments.

Our method consists of a pre-training stage and a text-to-parameter translation stage. In the pre-training stage, we first train an imitator to imitate the rendering behavior of the game engine to make the parameter searching pipeline end-to-end differentiable. We also pre-train a translator to translate the CLIP image embeddings of random game characters to their facial parameters. Then at the text-to-parameter translation stage, on one hand, we fine-tune the translator on un-seen CLIP text embeddings to predict continuous parameters given text prompt rather than images, on the other hand, discrete parameters are evolutionally searched. Finally, the game engine takes in the facial parameters and creates the in-game characters which correspond to the text prompt described, as shown in Fig~\ref{fig:teaser}. Objective evaluations and subjective evaluations both indicate our method outperforms other SOTA zero-shot text-to-3D methods.

Our contributions are summarized as follows:

1) We propose a novel text-to-parameter translation method for zero-shot in-game character auto-creation. To the best of our knowledge, we are the first to study text-driven character creation ready for game environments.

2) The proposed T2P can optimize both continuous and discrete parameters in a unified framework, unlike earlier methods giving up controlling difficult-to-learn discrete parameters.

3) The proposed text-driven auto-creation paradigm is flexible and friendly for users, and the predicted physically meaningful facial parameters enable players or game developers to further finetune the game character as they want.

\begin{figure*}
    \centering{\includegraphics[width=1.0\linewidth]{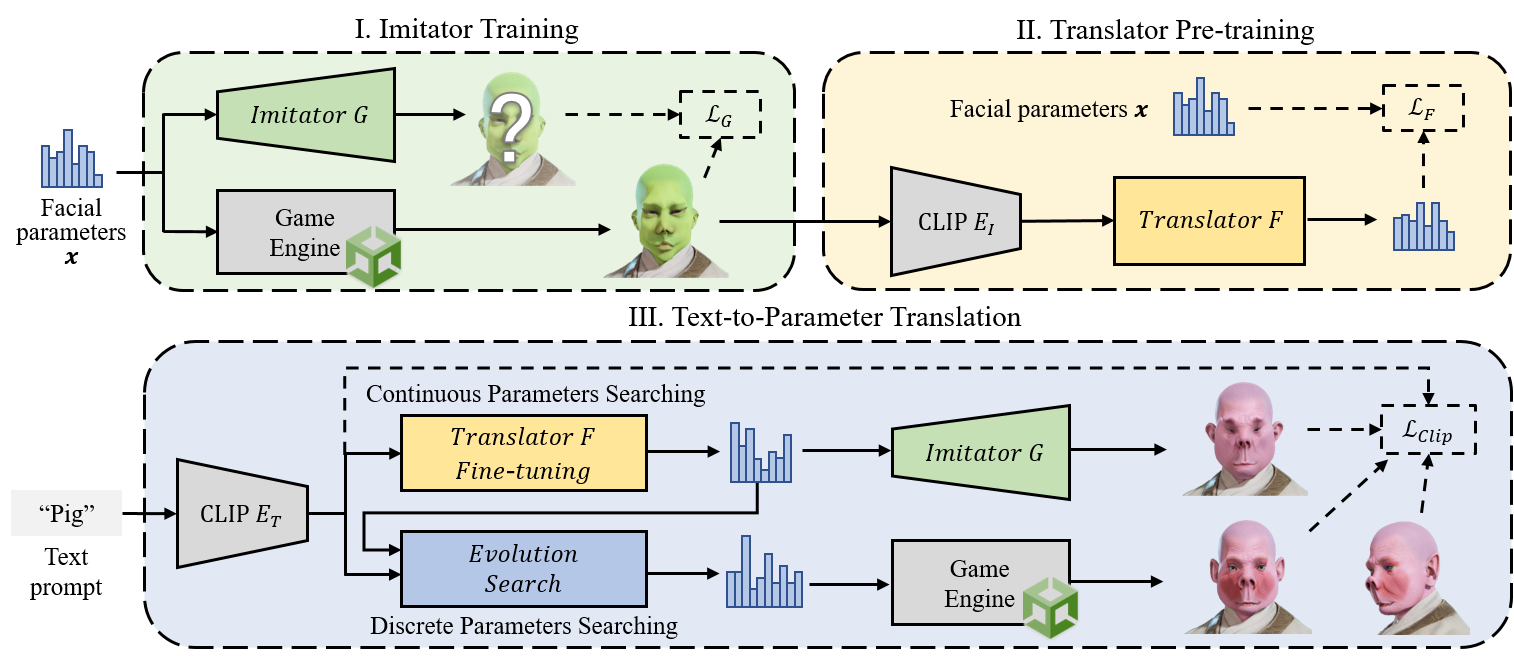}}\\
    \caption{An overview of the proposed T2P. $E_I$ and $E_T$ denote the CLIP image encoder and text encoder, respectively. An imitator is trained to mimic the game engine and achieve differentiable rendering. A translator is pre-trained to translate the CLIP image embeddings to continuous facial parameters. When creating game characters given text prompts, T2P searches continuous facial parameters by fine-tuning the translator and searches discrete facial parameters by the evolution search. Finally, the facial parameters are fed into the game engine to render the in-game characters.}
    \label{fig:overview}
\end{figure*}

\section{Related Work}
\label{sec:related work}
\subsection{Parametric Character Auto-Creation}
\label{sec:2.1}

Character auto-creation has been an emerging research topic because of its significance in role-playing games, augmented reality, and metaverses. Some methods on this topic are recently proposed. Tied Output Synthesis (TOS) learns to predict a set of binary facial parameters to control the graphical engine to generate a character that looks like the human in input photo~\cite{wolf2017unsupervised}. Face-to-Parameter translation (F2P) is proposed to optimize a set of continuous facial parameters to minimize the distance between the generated game character's face and the input photo~\cite{shi2019face}. 
In F2P's following works~\cite{shi2020fast, shi2020neural}, the framework is improved to achieve fast and robust character creation.
The PockerFace-Gan is proposed to decouple the expression features and identity features in order to generate expression-less game characters~\cite{shi2020neutral}. Borovikov et al. applies domain engineering and predict the facial parameters in a global-local way, considering the face as a hierarchical ensemble of general facial structure and local facial regions~\cite{borovikov2022applied}.
These methods all need reference photos to create characters, while we aim at creating characters based on text input.

\subsection{3D Face Reconstruction}

3D face reconstruction also aims to generate a 3D face given single or multi-view 2D facial images. 3D morphable model (3DMM)~\cite{blanz1999morphable} and its variants~\cite{cao2013facewarehouse, huber2016multiresolution, booth20173d, li2017learning, gerig2018morphable} are representative methods in the literature. They first parameterize a 3D face mesh data and then optimize it to match the facial identity, expression, and texture of given reference images. Taking advantage of deep Convolutional Neural Networks (CNNs), high-level image representations are used to improve the predicting of the morphable model coefficients~\cite{tuan2017regressing, dou2017end, jackson2017large}. The recently proposed MeInGame firstly reconstructs the face as a 3DMM model and then transfers the face to game mesh keeping their topology~\cite{lin2021meingame}. 
It also predicts texture map and lighting coefficients from input images to improve the outlook of the game mesh.

\subsection{Zero-Shot Text-Driven Generation}

Zero-shot content generation is recently made possible by the powerful multimodel representation and generalization capabilities of CLIP~\cite{reed2016generative}.
Combining the CLIP with variational autoencoder or diffusion model, DALL-E~\cite{ramesh2021zero}, DALL-E 2~\cite{ramesh2021zero} and Imagen~\cite{ramesh2022hierarchical} achieved high-quality zero-shot text-to-image synthesis, and sparked widespread discussion. Text-driven image translation and manipulation, and human image generation are also explored~\cite{patashnik2021styleclip, kim2021diffusionclip, gal2022stylegan, wei2022hairclip, jiang2021talk, xia2021tedigan, xu2022predict, yu2022towards, jiang2022text2human}. 
Taking advantage of CLIP, zero-shot text-driven 3D object generation and manipulation methods made rapid advances~
\cite{jain2022zero, michel2022text2mesh, sanghi2022clip,wang2022clip, canfes2022text, khalid2022clip}. 
The most recently proposed Dreamfusion uses Imagen to supervise the Neural Radiance Fields network (NeRF)~\cite{mildenhall2021nerf} to generate 3D object~\cite{poole2022dreamfusion}.
The most related work to ours named AvatarCLIP was recently proposed to achieve zero-shot text-driven 3D avatar generation and animation~\cite{hong2022avatarclip}. Given a text prompt, AvatarCLIP first generates a coarse shape by code-book-based retrieval, guided by CLIP. Then the coarse shape is used to initialize a NeuS network~\cite{wang2021neus} to generate the implicit representation. Finally, the implicit 3D avatar is optimized to sculpt fine geometry and generate texture.
This method treats the 3D human generation as a NeuS optimization process. However, the implicit representation makes it difficult to implement in games and unfriendly to user interaction. As a comparison, our created bone-driven game characters are controlled by explicit parameters with physical meanings. This enables players and game developers to further edit the created characters according to their needs.


\section{Method}
\label{sec:method}

Fig.~\ref{fig:overview} shows an overview of the proposed T2P. We first train an imitator to simulate the game engine and pre-train a translator to translate the CLIP image embeddings to continuous facial parameters. Then, to achieve text-to-parameter translation, given the text prompts, we fine-tune the translator to predict continuous parameters and combine the evolution search to optimize discrete parameters.

\subsection{Imitator}
\label{sec:imitator}

We train a neural imitator to mimic the behavior of the game engine in order to differentiate the rendering of in-game characters. It takes in continuous facial parameters $\bm{x}$ and renders the front view of the game character $\bm{y}$. Different from the F2P~\cite{shi2019face} taking a similar generator network architecture of DC-GAN~\cite{radford2015unsupervised}, we add a positional encoder at the input-end of the renderer to improve the facial parameters parsing on complex textures and geometry. We treat the imitator training as a regression problem to minimize the pixel-wise distance between the images rendered by the game engine and the imitator. To avoid the blurry rendered pixels, we use L1 loss as the loss function to train the imitator:
\begin{equation}
\begin{split}
   \mathcal{L}_{G}(\bm{x})
   &=E_{\bm{x} \sim u(\bm{x})}\{||\bm{y}-\hat{\bm{y}}||_1\}\\
   &=E_{\bm{x} \sim u(\bm{x})}\{||G(\bm{x})-Engine(\bm{x})||_1\},
\end{split}
\end{equation}
where $G(\bm{x})$ and $Engine(\bm{x})$ represent the image rendered by the imitator and game engine, respectively.

To prepare the training data, we randomly sample 170K continuous facial parameters $\bm{x}$ from a multidimensional uniform distribution $u(\bm{x})$. We feed these parameters into the game engine to render out the facial images. Then these facial parameters and image pairs are split into 80\% and 20\% for training and validation.

\subsection{Continuous Parameters Searching}

\begin{figure}
    \centering{\includegraphics[width=1.0\linewidth]{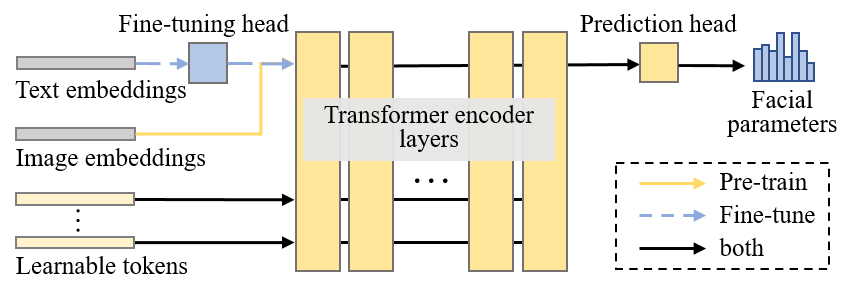}}\\
    \caption{The architecture of our translator. The translator contains a set of transformer encoder layers, several learnable tokens, a fine-tuning head, and a prediction head. The translator is firstly pre-trained on CLIP image embeddings and then fine-tuned on CLIP text embeddings to predict continuous facial parameters. When fine-tuning the translator, only the parameters of the fine-tuning head are updated.}
    \label{fig:translator}
\end{figure}

We aim to train a translator to predict continuous facial parameters based on CLIP text embeddings. To reduce the learning difficulty, we first pre-train the translator on CLIP image embeddings and then fine-tune it on text CLIP embeddings. The main reason is that text-parameter pairs are expensive to collect, while image-parameter pairs can be infinitely generated with the game engine.

We take the randomly sampled facial parameters and rendered image pairs mentioned in section~\ref{sec:imitator} as training data. The rendered images are fed into the CLIP image encoder to collect image embeddings. Then we build a translator $F$ based on a transformer encoder, and train it to map the image embeddings $\bm{e}_I$ into facial parameters $\bm{x}$, as shown in Fig.~\ref{fig:translator}. The object function is defined as the L1 reconstruction loss between the true facial parameters and the predicted ones:
\begin{equation}
\begin{split}
   \mathcal{L}_{F}(\bm{e}_I, \bm{x})
   =E_{\bm{e}_I \sim u(\bm{e}_I)}\{||F(\bm{e}_I)-\hat{\bm{x}}||_1\}.
\end{split}
\end{equation}

When T2P creates game characters given text prompts, there is no image embeddings available. 
Though the CLIP is trained to pull the text and image pairs close to each other in the embedding space, there are still gaps between the two modalities. We, therefore, fine-tune the translator to fit the input text embeddings. 
Inspired by the recent prompt tuning study \cite{zhou2022learning}, we fix the parameters of the transformer and fine-tune a tiny tuner head. The translator is trained to map the text embeddings $\bm{e}_T$ to facial parameters $\bm{x}$. Then the facial parameters are fed into the imitator to render the image of the game character. The fine-tuning object function is to minimize the cosine distance between the given text embeddings $\bm{e}_T$ and the image embeddings of the rendered image:
\begin{equation}
\begin{split}
   \mathcal{L}_{CLIP}(\bm{e}_T, \bm{x})
   &=1-cos(\bm{e}_T, E_I(G(\bm{x})))\\
   &=1-cos(\bm{e}_T, E_I(G(F(\bm{e}_T))),
\end{split}
\end{equation}
where $E_I$ is the CLIP image encoder. The parameters of the fine-tuned head $w$ are iteratively updated as follows,
\begin{equation}
\begin{split}
   w \leftarrow w - \eta_t \frac{\partial\mathcal{L}_{CLIP}}{\partial w},
\end{split}
\end{equation}
where $\eta_t$ is the learning rate at $t$th iteration. We follow the snapshot ensembles~\cite{huang2017snapshot} and set the learning rate using the cosine annealing schedule with warm restarts (SGDR)~\cite{loshchilov2016sgdr} to encourage the translator to converge to and escape from local minima:
\begin{equation}
    \eta_t = \eta_{min}+\frac{1}{2}(\eta_{max}-\eta_{min})(1+cos(\frac{N_t}{N}\pi)),
\end{equation}
where $\eta_{min}$, $\eta_{max}$, and $\eta_{t}$ denote the minimum, maximum, and current learning rate, respectively. $N$ denotes the number of iterations between two warm restarts, and $N_t$ denotes the number of iterations since the last restart. Each time the $N_t$ equals $N$, the current iteration is called a snapshot point, and we save the predicted facial parameters at this point. These facial parameters are then used to initialize the first population of the evolution search.

\begin{figure*}
    \centering{\includegraphics[width=0.96\linewidth]{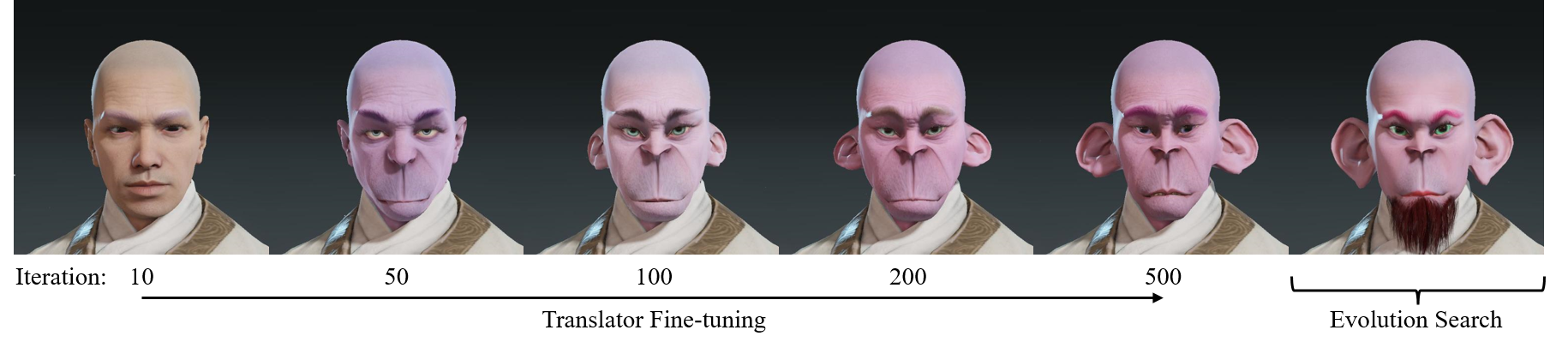}}\\
    \caption{Game characters created by the proposed T2P given the text prompt ``monkey''. The first five game characters are created by the translator at different fine-tuning iterations. The last one is created by the evolution search, adding a discrete facial element, a beard.}
    \label{fig:results_1}
\end{figure*}

\subsection{Discrete Parameters Searching}

In the bone-driven face model, besides continuous facial parameters controlling its bones, discrete facial elements (like the hairstyle, beard styles, and make-up) are also important. However, these elements are difficult for the imitator to learn, because they are discrete and highly changeable. Unlike previous methods that ignore discrete parameters during optimization, we propose to evolutionally search them by directly interacting with the game engine. Evolutionary algorithms have been widely used in reinforcement learning and neural architecture search~\cite{salimans2017evolution, liu2021survey}, where the objective function can be optimized without using any gradient information.

Here we perform a text-driven evolution search to find the optimum discrete facial parameters. The initial generation contains random initialized discrete parameters as well as the continuous facial parameters predicted by the translator. To impose supervision on 3D views, we render out two images for each game character, one for front view $\bm{y}_{front}$ and one for side view $\bm{y}_{side}$.
The facial parameters are scored by the CLIP model as follows,
\begin{equation}
\begin{split}
    S_{CLIP} &=\alpha cos(E_T(T), E_I(\bm{y}_{front}))\\
    & +(1-\alpha) cos(E_T(T'), E_I(\bm{y}_{side})),
\end{split}
\end{equation}
where $\alpha$ is the weight coefficient, $T$ is the given text prompt, $T'$ is the automatically adjusted text prompt for the side view, $E_T$ is the CLIP text encoder and $E_I$ is the CLIP image encoder.
Then $k$ random pairs of facial parameters are selected as parents to produce the next generation through crossover and mutation. For the crossover step,  
child $\bm{x}^c$ is generated by randomly choosing a value from parents $\bm{x}^f$ and $\bm{x}^m$ at each position $i$,
\begin{equation}
   P(x^c_i=x^f_i)+P(x^c_i=x^m_i)=1.
\end{equation}
For the mutation step, each child parameter $\bm{x}^c$ is added random noise at multiple randomly selected position $i$,
\begin{equation}
    {x^c_i}'=x^c_i+noise.
\end{equation}
The newly generated children's parameters together with the better ones of the parents' parameters are selected as the next generation and get involved in the looping selection, crossover, and mutation. The evolution process terminates until the CLIP score is converged.

\subsection{Implementation Details}

\textbf{Network architecture.} Our imitator consists of a positional encoder with four fully-connected layers and a generator with six transposed convolution layers. The generator is similar to DCGAN's generator~\cite{radford2015unsupervised}, except that its Tanh activation of the output layer is removed to encourage a better convergence. The translator consists of eight Transformer encoder layers~\cite{vaswani2017attention}, each of them having eight multi-attention heads, and sixteen input tokens. The first token is the CLIP embeddings and the other tokens are learnable. We concatenate a prediction head with one single fully-connected layer after the Transformer. The fine-tuning head of the translator is a three layers perceptron with a bottleneck architecture.

\textbf{Training details.} The imitator and translator are both trained using SGD optimizer~\cite{bottou2010large}. We set the momentum to 0.9 and set the weight decay to 5e-4. For imitator pretraining, the learning rate is set to 1e-3 and is reduced to 0.98x per 30 epochs, and the training is stopped after 500 epochs. For translator pre-training, the learning rate is set to 1e-4 and is reduced to 0.1x at the 600th epoch and the training is stopped at the 1000th epoch. We randomly sample 170K facial parameters and corresponding rendered images of in-game characters pairs to train the imitator and translator. For translator fine-tuning, the minimum and maximum learning rates are set to $\eta_{min}=0$ and $\eta_{max}=1$, respectively, and the number of iterations between two warm starts $N$ is set to 10 for the SGDR learning rate scheduler. Fine-tuning is stopped when the CLIP scores are no longer improved by more than 100 iterations.

\textbf{Evolution search.} The facial parameters predicted by the translator at the last 5 snapshot points are selected as initial values. Each set of facial parameters contains 269 continuous parameters and 62 discrete parameters, and the initialized values of these discrete parameters are set to zeros, which means these facial elements do not appear at the beginning. These 5 sets of facial parameters together with 5 more random ones are the first population for the evolution search. We found that updating continuous parameters together with discrete parameters in the evolution search achieves better results. The number of selected pairs of parents is set to 10. The weight coefficient $\alpha$ is set to 0.8. The crossover rate is set to 0.4 and the mutation rate is set to 0.05. 

\textbf{Prompt engineering.} To enhance the text prompts, we follow the CLIP~\cite{radford2021learning} and adapt prompt ensembling to the given text prompts. We preset 12 template sentences, such as ``\{\} head rendered in a game engine'', and then fill the ``\{\}'' with the input text prompt. We calculate the CLIP text embeddings of the filled sentences and take their mean value as the input text embeddings for the translator and evolution search. For evolution search, we further add ``side view of'' to the template sentences when calculating the CLIP score of the rendered images of the side view.


\section{Experimental Results and Analysis}
\label{sec:experments}

\begin{figure}
    \centering{\includegraphics[width=0.96\linewidth]{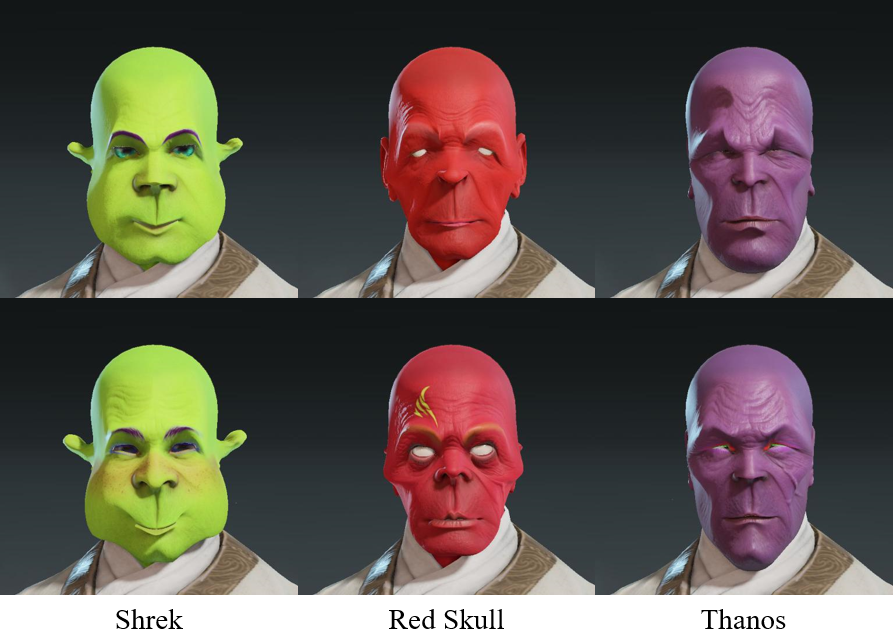}}\\
    \caption{In-game fictional characters created by the proposed T2P given different text prompts. The results in the first row are created by the translator. The results in the second row are created by the evolution search.}
    \label{fig:results_2}
\end{figure}

\begin{figure*}
    \centering{\includegraphics[width=0.99\linewidth]{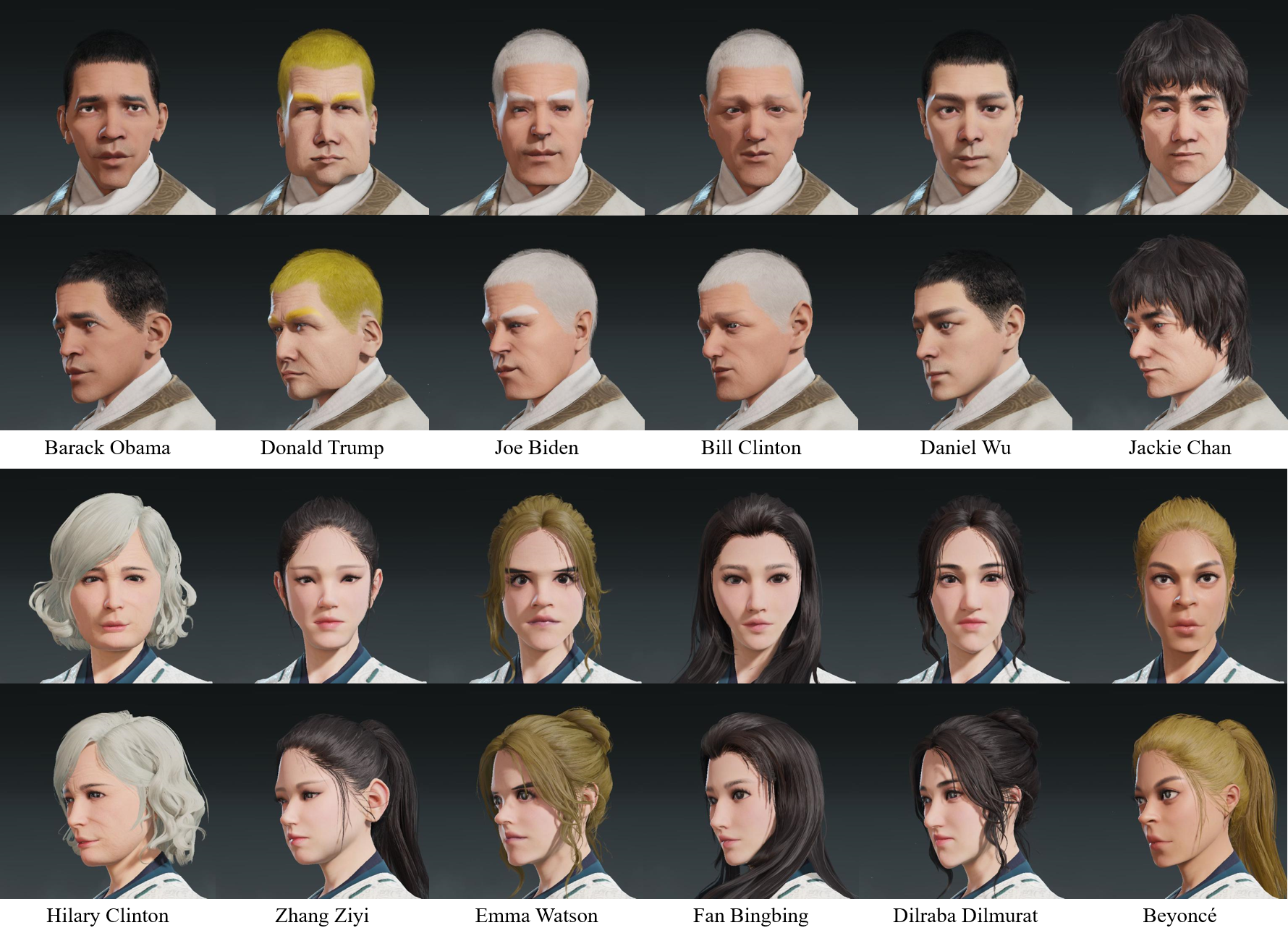}}\\
    \caption{In-game celebrities created by the proposed T2P. This figure shows the front view and the side view for each character.}
    \label{fig:results_3}
\end{figure*}

\begin{figure}
    \centering{\includegraphics[width=1.0\linewidth]{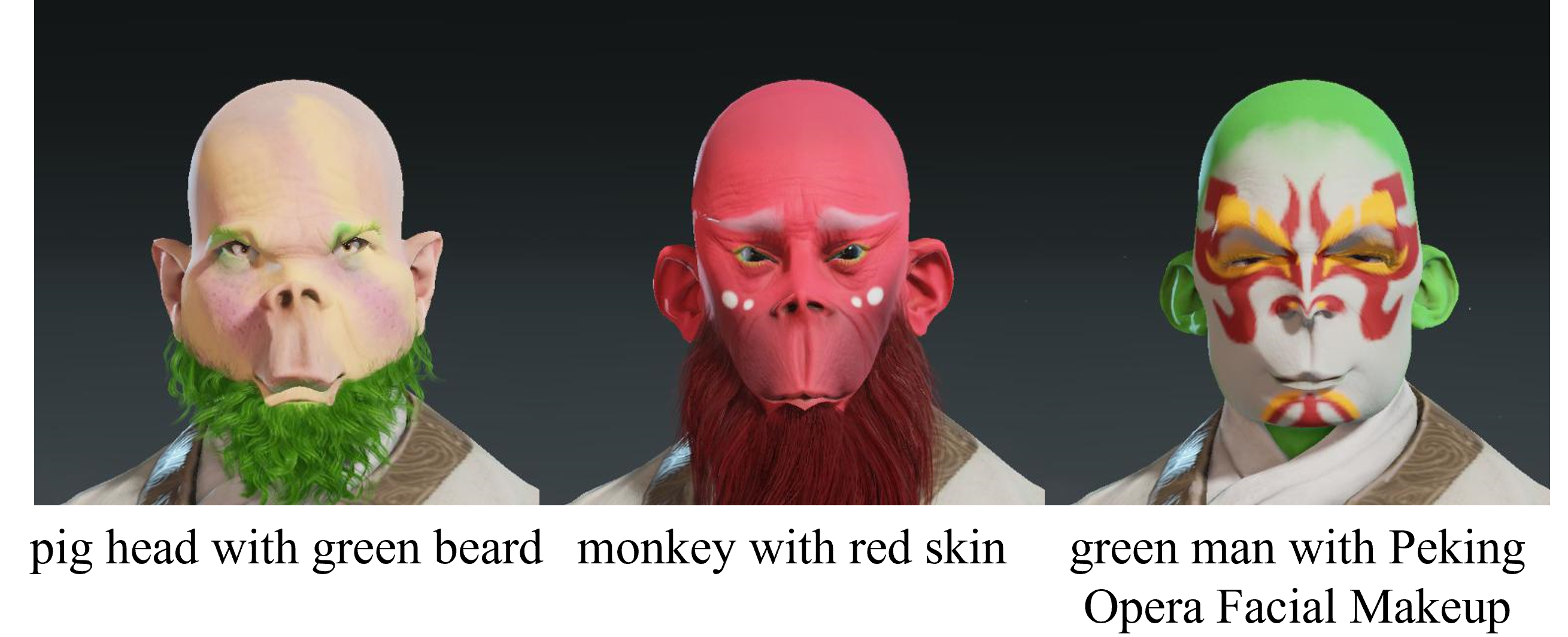}}\\
    \caption{In-game characters created by the proposed T2P given complicated prompts.}
    \label{fig:long_prompt}
\end{figure}

\subsection{Game Character Auto-Creation}

Fig.~\ref{fig:results_1} shows the game characters created by T2P given the text prompt ``monkey''. The first five images show the in-game characters created by the translator at different fine-tuning iterations. The in-game character gradually grows from a normal human face to look like a monkey. The evolution search further searches discrete facial elements and also slightly improves continuous parameters. The last image of Fig.~\ref{fig:results_1} shows the evolution search adds a beard to the character to make it more vivid. 
In this process, the proposed T2P is enabled to search both continuous and discrete facial parameters to optimize the in-game character to be consistent with the given text prompt and vivid. 
Fig.~\ref{fig:results_2} shows more results of fictional character creation. Results in the first row are controlled by continuous parameters, and results in the second row are added discrete facial elements.

T2P can create characters with animal heads, as shown in Fig.~\ref{fig:results_1}, fictional characters, as shown in Fig.~\ref{fig:results_2}, and celebrities, as shown in 
Fig.~\ref{fig:results_3}, and characters conditioned on complactied text prompts, as shown in Fig.~\ref{fig:long_prompt}. 
These results show the powerful zero-shot game character auto-creation ability of the proposed T2P. By inputting only a text prompt, T2P can generate a vivid character, which is more flexible and time-saving for players or game developers compared to manual customization.

\begin{table*}
  \centering
  \begin{tabular}{c|ccc|cc}
    \toprule
	\multirow{2}{*}{Method} & \multicolumn{3}{c}{Objective Evaluations} & \multicolumn{2}{c}{Subjective Evaluations}\\
     & Inception Score $\uparrow$& CLIP Ranking-1 $\uparrow$ & Time $\downarrow$ & Reality $\uparrow$ & Consistency with Text $\uparrow$\\
    \midrule
    DreamFusion~\cite{poole2022dreamfusion} &$1.60 \pm 0.12$&$16.67\%$&$254.50$min& $1.85 \pm 1.02$ &$2.23 \pm 1.39$\\
    AvatarCLIP~\cite{hong2022avatarclip} &$1.37 \pm 0.31$&$16.67\%$&$177.79$min& $1.97 \pm 0.53$ &$2.14 \pm 0.66$\\
    T2P (ours) &$\bm{1.65} \pm 0.21$&$\bm{66.66\%}$&$\bm{359.47}$s& $\bm{3.87} \pm \bm{0.47}$ &$\bm{3.34} \pm \bm{0.53}$\\
    \bottomrule
  \end{tabular}
  \caption{Comparison results of DreamFusion, AvatarCLIP, and the proposed T2P in terms of objective and subjective evaluations.}
  \label{tab:comparison_results}
\end{table*}

\subsection{Comparison with Other Methods}

We compare the proposed method with 
AvatarCLIP~\cite{hong2022avatarclip} and DreamFusion~\cite{poole2022dreamfusion}. The comparison includes objective evaluations and subjective evaluations. 
Since DreamFusion is not open source yet, we use the community implementation version of it, named Stable-Dreamfusion\footnote[1]{https://github.com/ashawkey/stable-dreamfusion}. This version uses the open-source stable diffusion model~\cite{rombach2022high} to drive the 3D object generation. 
We only compare the heads generated by these methods. This may introduce unfairness, thus we will never claim superiority besides the head part.

\begin{figure}
    \centering{\includegraphics[width=1.0\linewidth]{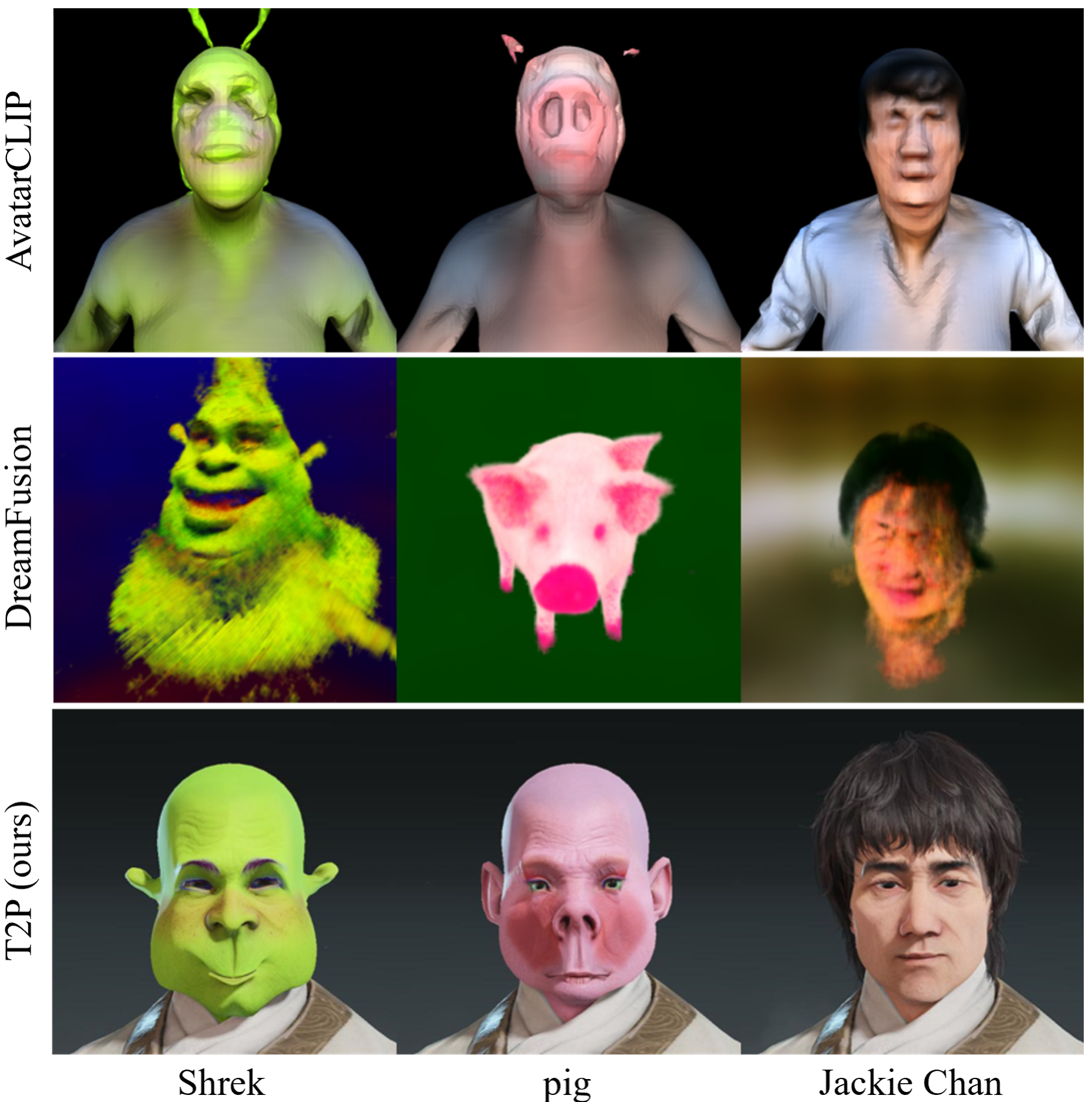}}\\
    \caption{Comparison of AvatarCLIP, DreamFusion, and the proposed T2P. Each column shows the 3D characters created by these methods given the same text prompt.}
    \label{fig:user_study_show}
\end{figure}

We feed 24 different text prompts into these two methods and our proposed T2P to generate characters respectively. Three examples are shown in Fig.~\ref{fig:user_study_show}. For objective evaluations, we compare the Inception Score~\cite{salimans2016improved}, CLIP Ranking-1, and their speed (run on NVIDIA A30), as shown in Table~\ref{tab:comparison_results}. For each method, CLIP Ranking-1 calculates the ratio of its created characters ranked by CLIP as top-1 among the characters created by all three methods. The evaluation scores show the proposed T2P outperforms the other two methods and runs at a much faster speed. 

For subjective evaluations, we invite 20 volunteers to evaluate the generation results in terms of realistic degree and consistency with the given text. They are asked to focus on the heads and faces of the characters and score them from 1 to 5, where 1 is the worst and 5 is the best. The evaluation results are shown in Table~\ref{tab:comparison_results}. Evaluation results show our method consistently outperforms the other two methods. We also notice that AvatarCLIP performs good at celebrities generation, Dreamfusion is good at fictional characters generation, while our method performs better at both types, just as shown in Fig.~\ref{fig:user_study_show}.

\subsection{Ablation Studies}

\begin{figure}
    \centering{\includegraphics[width=1.0\linewidth]{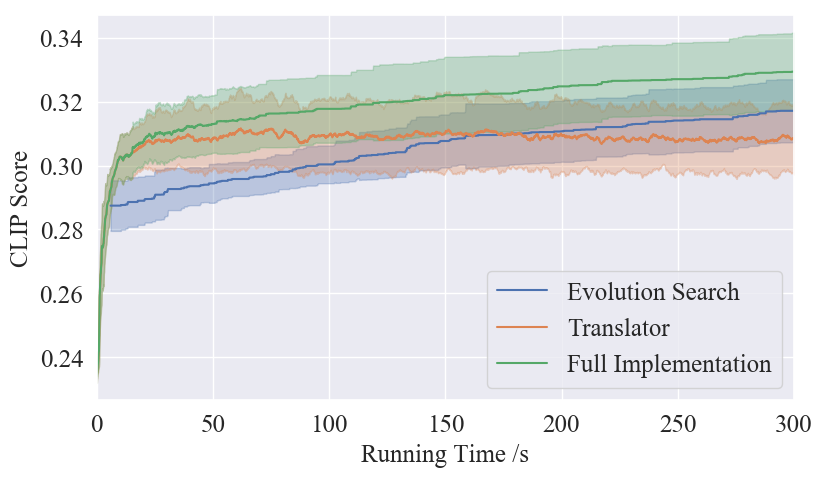}}\\
    \caption{Curves of CLIP scores increasing within 300s under three different module settings.}
    \label{fig:ablation_studies}
\end{figure}

We conduct ablation studies to analyze the importance of the proposed translator and evolution search. We run our framework with three settings, including 1) only evolution search 2) only translator and 3) both translator and evolution search. The details of these settings are as follows.

1) \textbf{Evolution Search.} The translator is removed from the framework and the evolution search is used to directly search both continuous and discrete facial parameters given text prompts.

2) \textbf{Translator.} The evolution search is abandoned, and the translator is fine-tuned to translate the given text prompts into continuous facial parameters and gives up controlling discrete parameters.

3) \textbf{Full Implementation.} Given text prompts, the translator is fine-tuned to predict continuous facial parameters. Then, the evolution search further searches discrete parameters and also improves the continuous ones.

Fig.~\ref{fig:ablation_studies} shows the CLIP scores increasing curves with the T2P running in 300 seconds. The means and standard deviations are calculated based on 100 times repeat running driven by one text prompt. As shown in the figure, the full implementation of our method always outperforms the other two. The translator is optimized rapidly to find optimal continuous parameters but can not further improve the CLIP scores because of lacking discrete facial elements. Compared with the translator, the evolution search is quite slow but can reach a higher CLIP score. 
The full implementation of T2P takes advantage of both translator and evolution search and achieves fast and better optimization.

We further test different settings of proposed T2P on 100 different text prompts to evaluate their performance. Table~\ref{tab:ablation_stydies} shows the results. The first row is the result of directly using the pre-trained translator to predict continuous facial parameters, and the second row is the result of fine-tuning translator to predict parameters. The fine-tuned one can achieve a higher CLIP score, which indicates the necessity of fine-tuning. The CLIP scores of only using the evolution search and the full version of T2P are shown in the third and fourth rows, respectively. The full version of T2P achieves the highest CLIP score because it can search both continuous and discrete facial parameters to create better in-game characters.

\begin{table}
  \centering
  \begin{tabular}{cc|c}
    \toprule
    Translator & Evolution Search & CLIP Score\\
    \midrule
    fixed & $\times$ &$27.29 \pm 3.10$\\
    fine-tuned & $\times$ &$34.85 \pm 3.15$\\
    $\times$ & $\checkmark$ &$35.31 \pm 2.26$\\
    fine-tuned & $\checkmark$ &$35.72 \pm 2.70$\\    
    \bottomrule
  \end{tabular}
  \caption{Results of ablation studies. Four versions of the proposed method are compared.}
  \label{tab:ablation_stydies}
\end{table}

\subsection{Facial Paramter Interpolation}

\begin{figure}
    \centering{\includegraphics[width=1.0\linewidth]{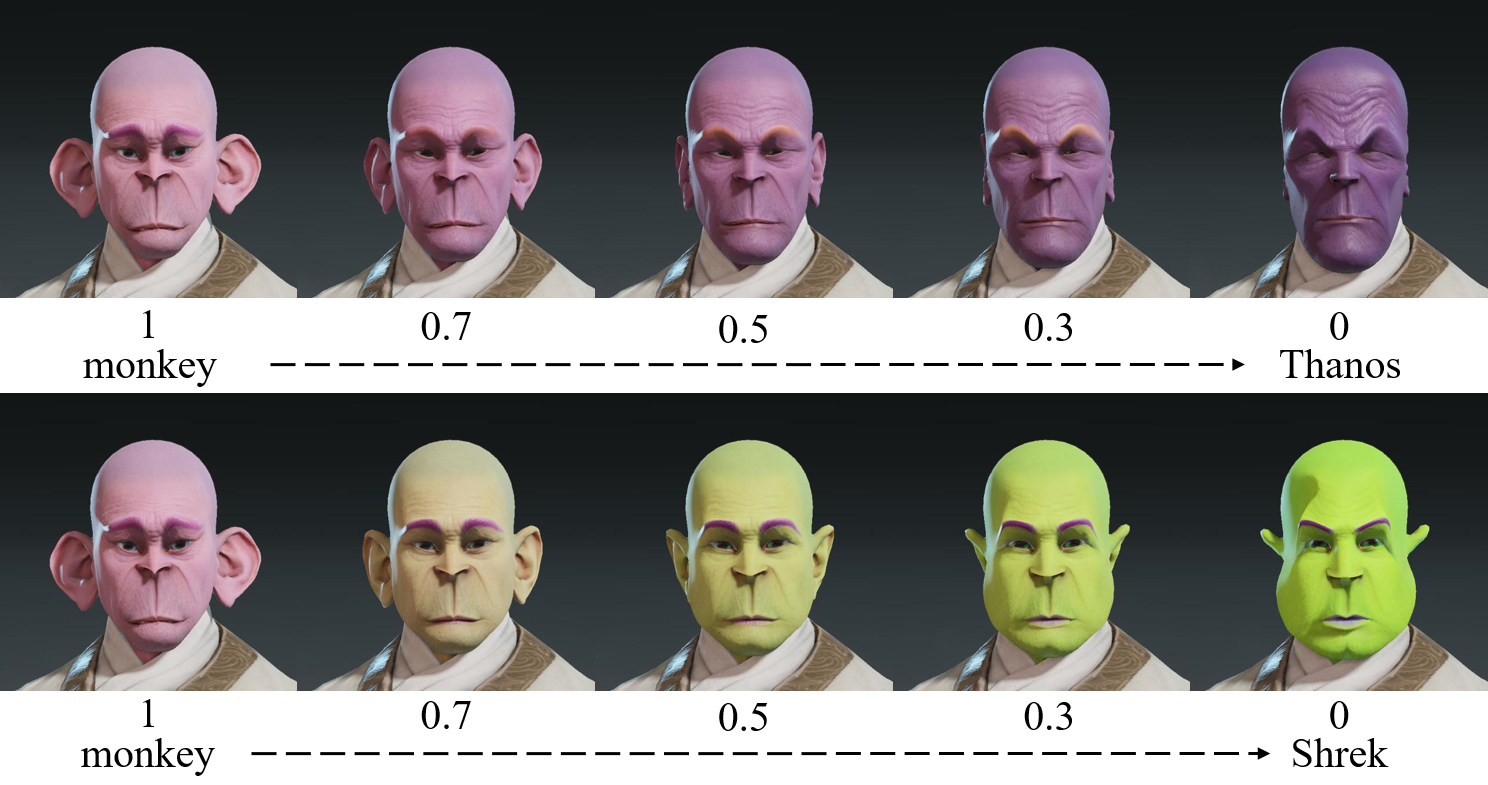}}\\
    \caption{Examples of the facial parameter interpolation of game characters.
    }
    \label{fig:results_4}
\end{figure}

Since the generated characters are controlled by parameters with explicit physical meanings, users can further adjust the outlook of the characters as they want.
One can also interpolate different facial parameters to create a new character, as shown in Fig.~\ref{fig:results_4}. The first row shows the interpolation between the monkey and Thanos, in which the new facial parameters are calculated as follows,
\begin{equation}
    \bm{x}_{new} =\beta \bm{x}_{monkey} +(1-\beta) \bm{x}_{Thanos},
\end{equation}
where $\beta$ is the interpolation coefficient decreasing from $1$ to $0$. The results in the second row of Fig.~\ref{fig:results_4} show the interpolation between the monkey and Shrek. Besides, more than two characters can also be interpolated. We believe the benefits of the facial parameters controlling bone-driven game characters can give players a higher degree of freedom in character customization.

\section{Conclusion}
\label{sec:conclusion}

We propose a novel method called ``text-to-parameter translation'' to create bone-driven in-game characters given text prompts. Our method achieves high-quality zero-shot creation of in-game characters and can search both continuous and discrete facial parameters in a unified framework. The proposed text-driven framework is flexible and time-saving for users, and the created bone-driven characters with physically meaningful facial parameters are convenient for users to further edit as they want. Experimental results show our method achieves high-quality and vivid zero-shot text-driven game character auto-creation and outperforms other SOTA text-to-3D generation methods in terms of objective evaluations, speed, and subjective evaluations.


{\small
\bibliographystyle{ieee_fullname}
\bibliography{egbib}
}

\clearpage



\appendix
\section*{Appendix}

\section{Module Details}
\subsection{Imitator}
\begin{figure}
    \centering{\includegraphics[width=1.0\linewidth]{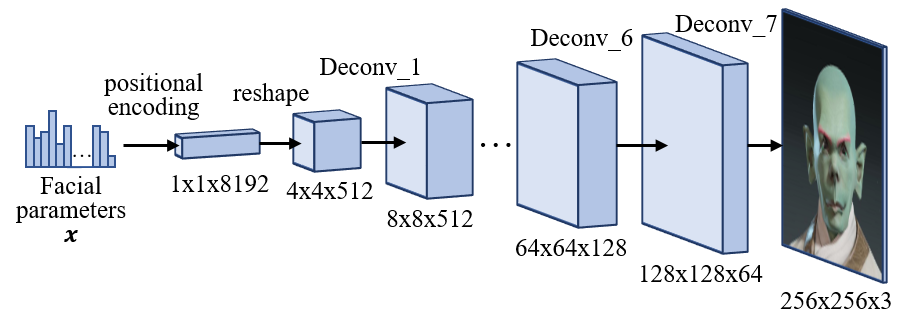}}\\
    \caption{The architecture of the imitator. We train the imitator to take in the continuous facial parameters $\bm{x}$ and render the facial image of the corresponding in-game character.}
    \label{fig:imitator_architecture}
\end{figure}

Fig.~\ref{fig:imitator_architecture} demonstrates the architecture of the proposed imitator. Table~\ref{tab:config_imitator} shows the detailed configurations of the imitator. The proposed imitator consists of a positional encoder and a generator. The positional encoder is built with four fully connected layers, and the generator is built with deconvolution layers, batch normalization layers, and ReLU activation layers. The imitator is trained to map the continuous facial parameters $\bm{x}$ to the rendered facial image of the in-game character.

\begin{table*}[]
\centering
\begin{tabular}{@{}l|ccccc@{}}
\toprule
                           & Module                              & Layer     & Component                 & Configuration & Output Size \\ \midrule
\multirow{11}{*}{Imitator} & \multirow{4}{*}{Positional Encoder} & FC\_1     & Fully connected layer     & 269-1024      & 1x1x1024    \\
                           &                                     & FC\_2     & Fully connected layer     & 1024-2048     & 1x1x2048    \\
                           &                                     & FC\_3     & Fully connected layer     & 2048-4096     & 1x1x4096    \\
                           &                                     & FC\_4     & Fully connected layer     & 4096-8192     & 1x1x8192    \\ \cmidrule(l){2-6} 
                           & \multirow{7}{*}{Generator}          & Deconv\_1 & Deconvolution + BN + ReLU & 512x4x4/2     & 4x4x512     \\
                           &                                     & Deconv\_2 & Deconvolution + BN + ReLU & 512x4x4/2     & 8x8x512     \\
                           &                                     & Deconv\_3 & Deconvolution + BN + ReLU & 512x4x4/2     & 16x16x512   \\
                           &                                     & Deconv\_4 & Deconvolution + BN + ReLU & 256x4x4/2     & 32x32x256   \\
                           &                                     & Deconv\_5 & Deconvolution + BN + ReLU & 128x4x4/2     & 64x64x128   \\
                           &                                     & Deconv\_6 & Deconvolution + BN + ReLU & 64x4x4/2      & 128x128x64  \\
                           &                                     & Deconv\_7 & Deconvolution             & 3x4x4/2       & 256x256x3   \\ \cmidrule(l){1-6} 
\end{tabular}
\caption{Detailed configurations of the imitator. For the configuration, in the $x$-$y$ of the fully connected layer, $x$ denotes the number of input nodes, and $y$ denotes the number of output nodes. In the $c \times h \times w /s$ of the Deconvolution layer, $c$ denotes the number of filters, $h \times w$ denotes the filter's size, and $s$ denotes the stride. For the output size, in the $ h \times w \times c$, $h \times w$ denotes the spatial size, and $c$ denotes the number of channels.}
\label{tab:config_imitator}
\end{table*}

\subsection{Evolution Search}
\begin{figure}
    \centering{\includegraphics[width=1.0\linewidth]{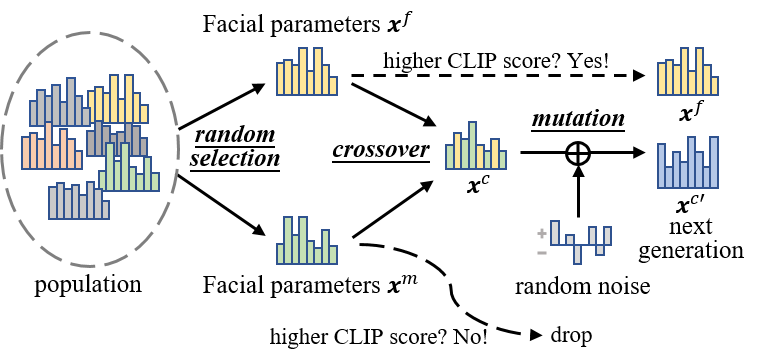}}\\
    \caption{Illustration of the text-driven evolution search. For the new generation, individuals are generated by selection, crossover, and mutation.}
    \label{fig:evolution_search}
\end{figure}

Fig.~\ref{fig:evolution_search} is an illustration of the text-driven evolution search. For the crossover, two sets of facial parameters $\bm{x}^f$ and $\bm{x}^m$ are randomly selected from the population as parents, then the child parameters $\bm{x}^c$ are generated by randomly selecting values from $\bm{x}^f$ or $\bm{x}^m$ at each position. For the mutation, random noises are added to the child parameters $\bm{x}^c$ at randomly selected positions. Then the child and the parent with a higher CLIP score are selected as two of the individuals of the next generation, while the parent with a lower CLIP score is dropped. The CLIP score of the best individual in each generation increases gradually through the looping of selection, crossover, and mutation.


\section{Continuous and Discrete Facial Parameters}
The bone-driven in-game character is controlled by continuous and discrete parameters, as shown in Table~\ref{tab:discrete_parameters} and Table~\ref{tab:continuous_parameters}. In these tables, the ``Primary Group'' represents the main parts of the face model, the ``Secondary Group'' represents the subparts of each primary facial part, the ``Controllers'' denotes the user-adjustable facial parameters and the ``\# Controllers'' denotes the number of controllers in each secondary group. The continuous facial parameters control the translation, rotation, and scale of facial bones, and the outlooks of some facial elements. The total number of continuous facial parameters is 269. The discrete facial parameters mainly control the outlooks of some discrete facial elements. Since some facial elements are difficult for the imitator to learn, besides their types, we also treat their translation, rotation, scale, and so on as discrete parameters. The total number of discrete facial parameters is 62.

\begin{table*}[]
\centering
\begin{tabular}{@{}l|l|l|l|c@{}}
\toprule
                                                                           & Primary Group             & Secondary Group       & Controllers                                                 & \# Controllers        \\ \midrule
\multirow{15}{*}{\rotatebox{90}{Discrete Parameters}}   & Eyebrows                  & Eyebrows Style        & type                                                                                                                                                                                                             & 1                    \\ \cmidrule(l){2-5} 
                                                                           & \multirow{4}{*}{Eyes}     & Eyeball            & type & \multirow{4}{*}{20}  \\
                                                                               &                           & Eyelids               & type                                                                                                                                                                                                             &                      \\
                                                             &                           & Eyelashes             & type, scale (whole), color (R, G, B)                                                                                                                                                                             &                      \\

                                                                                                                                             &                           & Eye Makeup               & \begin{tabular}[c]{@{}l@{}}type of eye makeup, eyeliner shading, eyeliner color (R, G, B)\\ upper eyeshadow shading, upper eyeshadow color (R, G, B)\\ lower eyeshadow shading, lower eyeshadow color (R, G, B)\end{tabular}                                                                                                                                                                                                            &                      \\
                                                                                                                                             \cmidrule(l){2-5} 
                                                                           & Lip                       & Lip Makeup            & type, shading, luster, color (R, G, B)                                                                                                                                                                           & 6                    \\ \cmidrule(l){2-5} 
                                                                           & \multirow{5}{*}{Face}     & Blush                 & type, translation (x, y), scale (whole), color (R, G, B)                                                                                                                                                         & \multirow{5}{*}{31}  \\
                                                                           &                           & Tattoos               & type, translation (x, y), scale (x, y), shading, color (R, G, B)                                                                                                                                                          &                      \\
                                                                           &                           & Scars                 & type, translation (x, y), scale (whole), shading                                                                                                                                                                 &                      \\
                                                                           &                           & Beard (Upper)         & type, shading, color (R, G, B)                                                                                                                                                                                   &                      \\
                                                                           &                           & Beard (Lower)         & type, shading, color (R, G, B)                                                                                                                                                                                   &                      \\ \cmidrule(l){2-5} 
                                                                           & Hair                      & Hair Style            & type, color (R, G, B)                                                                                                                                                                                            & 4                    \\ \bottomrule
\end{tabular}
    \caption{Details of the discrete facial parameters. The x and y denotes the horizontal and vertical directions, respectively.}
    \label{tab:discrete_parameters}
\end{table*}

\begin{table*}[]
\centering
\begin{tabular}{@{}l|l|l|l|c@{}}
\toprule
                                                                           & Primary Group             & Secondary Group       & Controllers                                                 & \# Controllers        \\ \midrule
\multirow{39}{*}{\rotatebox{90}{Continuous Parameters}} & \multirow{4}{*}{Eyebrows} & Eyebrows (Inner)      & translation (x, y, z), rotation (y, z), scale (x, y, z)     & \multirow{4}{*}{29}  \\
                                                                           &                           & Eyebrows (Middle)     & translation (x, y, z), rotation (y, z), scale (x, y, z)     &                      \\
                                                                           &                           & Eyebrows (Outer)      & translation (x, y, z), rotation (y, z), scale (x, y, z)     &                      \\
                                                                           &                           & Outlook               & shading, density, color (R, G, B)                           &                      \\ \cmidrule(l){2-5} 
                                                                           & \multirow{7}{*}{Eyes}     & Eyes (Whole)          & translation (x, y, z), rotation (x, y, z), scale (whole)    & \multirow{7}{*}{58}  \\
                                                                           &                           & Upper Eyelids (Inner) & translation (x, y, z), rotation (x, y, z), scale (x, y, z)  &                      \\
                                                                           &                           & Upper Eyelids (Outer) & translation (x, y, z), rotation (x, y, z), scale (x, y, z)  &                      \\
                                                                           &                           & Lower Eyelids         & translation (x, y, z), rotation (x, y, z), scale (x, y, z)  &                      \\
                                                                           &                           & Inner Corner          & translation (x, y, z), rotation (x, y, z), scale (x, y, z)  &                      \\
                                                                           &                           & Outer Corner          & translation (x, y, z), rotation (x, y, z), scale (x, y, z)  &                      \\
                                                                           &                           & Eyeballs              & scale (whole), brightness, scale of pupils, color (R, G, B) &                      \\ \cmidrule(l){2-5} 
                                                                           & \multirow{6}{*}{Nose}     & Nose (Whole)          & translation (y, z), rotation (x)                            & \multirow{6}{*}{36}  \\
                                                                           &                           & Nose Bridge           & translation (y, z), rotation (x), scale (x, y, z)           &                      \\
                                                                           &                           & Septum                & translation (y, z), rotation (x), scale (x, y, z)           &                      \\
                                                                           &                           & Nostrils              & translation (x, y, z), rotation (x, y, z), scale (x, y, z)  &                      \\
                                                                           &                           & Nose Tip              & translation (y, z), rotation (x), scale (x, y, z)           &                      \\
                                                                           &                           & Base of Nose          & translation (y, z), rotation (x), scale (x, y, z)           &                      \\ \cmidrule(l){2-5} 
                                                                           & \multirow{6}{*}{Mouth}    & Mouth (Whole)         & translation (y, z), rotation (x)                            & \multirow{6}{*}{42}  \\
                                                                           &                           & Upper Lip (Middle)    & translation (y, z), rotation (x), scale (x, y, z)           &                      \\
                                                                           &                           & Upper Lip (Sides)     & translation (x, y, z), rotation (x, y, z), scale (x, y, z)  &                      \\
                                                                           &                           & Lower Lip (Middle)    & translation (y, z), rotation (x), scale (x, y, z)           &                      \\
                                                                           &                           & Lower Lip (Sides)     & translation (x, y, z), rotation (x, y, z), scale (x, y, z)  &                      \\
                                                                           &                           & Corners of Lips       & translation (x, y, z), rotation (x, y, z), scale (x, y, z)  &                      \\ \cmidrule(l){2-5} 
                                                                           & \multirow{12}{*}{Face}    & Forehead (Middle)     & translation (y, z), rotation (x), scale (x, y, z)           & \multirow{12}{*}{82} \\
                                                                           &                           & Brow Bone             & translation (y, z), rotation (x), scale (x, y, z)           &                      \\
                                                                           &                           & Forehead (Sides)      & translation (x, y, z), rotation (x, y, z), scale (x, y, z)  &                      \\
                                                                           &                           & Cheekbone             & translation (x, y, z), rotation (x), scale (x, z)           &                      \\
                                                                           &                           & Cheek (Upper)         & translation (x, y, z), rotation (x), scale (z)              &                      \\
                                                                           &                           & Cheek (Middle)        & translation (x, y, z), rotation (x, y), scale (z)           &                      \\
                                                                           &                           & Cheek (Lower)         & translation (x, y, z), rotation (x), scale (z)              &                      \\
                                                                           &                           & Philtrum (Sides)      & translation (x, y, z), rotation (x, y), scale (z)           &                      \\
                                                                           &                           & Chin (Middle)         & translation (x, y, z), rotation (x, y), scale (z)           &                      \\
                                                                           &                           & Chin (Sides)          & translation (x, y, z), rotation (x, y, z), scale (x, y, z)  &                      \\
                                                                           &                           & Mandible (Middle)     & translation (x, y, z), rotation (x, y, z), scale (x, y, z)  &                      \\
                                                                           &                           & Mandible (Sides)      & translation (x, y, z), rotation (x, y, z), scale (x, y, z)  &                      \\ \cmidrule(l){2-5} 
                                                                           & \multirow{3}{*}{Ears}     & Ears (Whole)          & translation (x, y, z), rotation (x, y, z), scale-whole      & \multirow{3}{*}{16}  \\
                                                                           &                           & Auricle               & translation (x, y, z), rotation (x)                         &                      \\
                                                                           &                           & Earlobe               & translation (x, y, z), rotation (x), scale (x)              &                      \\ \cmidrule(l){2-5} 
                                                                           & Skin                      & Skin                  & color (R, G, B), luster, aging, metallic                    & 6                    \\ \midrule
\end{tabular}
    \caption{Details of the continuous facial parameters. The x, y, z denotes the horizontal, vertical, and depth directions, respectively.}
    \label{tab:continuous_parameters}
\end{table*}

\section{Pre-training Samples}
\label{sec:train_samples}
\begin{figure*}
    \centering{\includegraphics[width=1.0\linewidth]{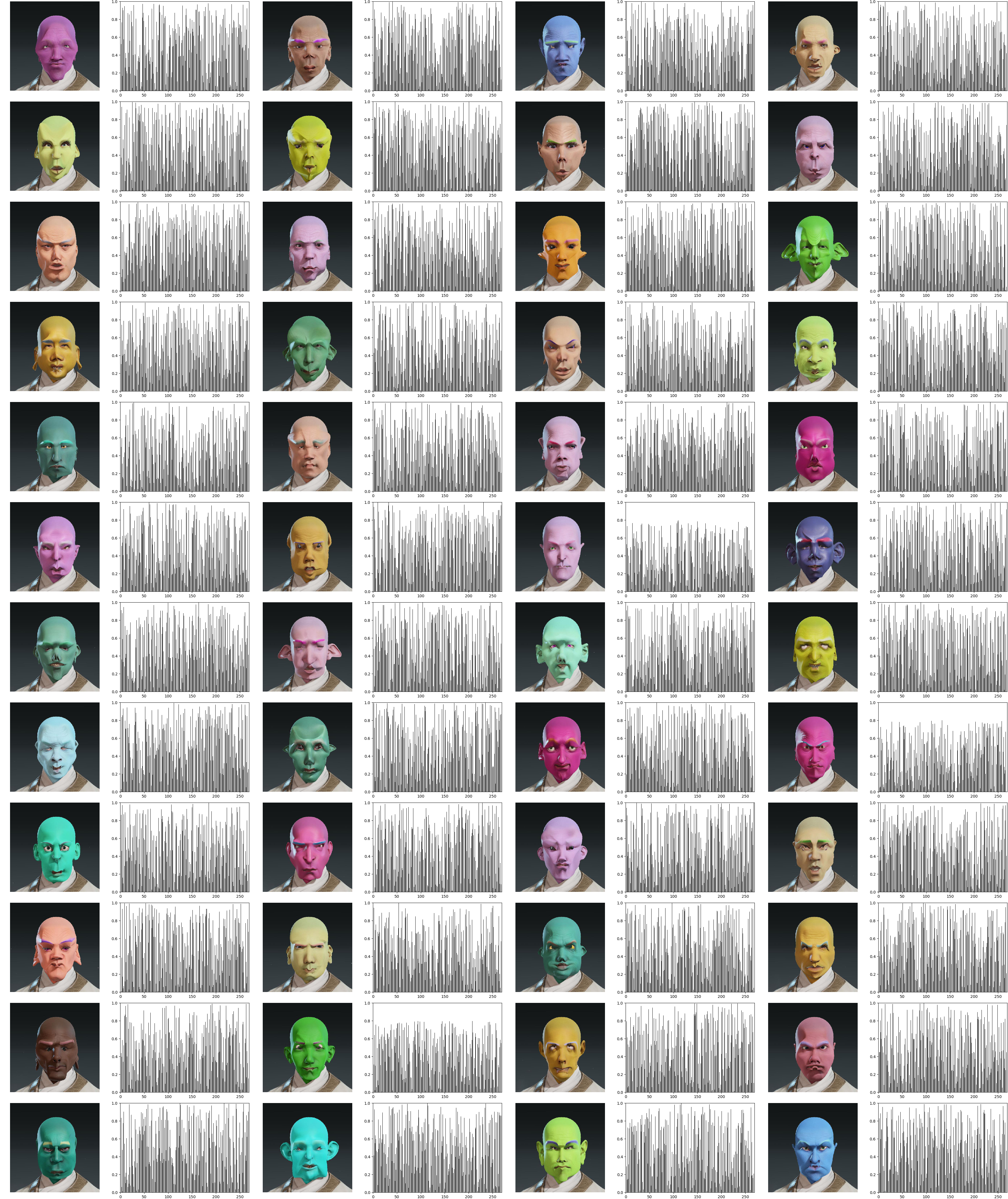}}\\
    \caption{Random sampled facial images of in-game characters and their corresponding continuous facial parameters.}
    \label{fig:training_samples}
\end{figure*}

We randomly sample 170,000 pairs of continuous facial parameters and rendered facial images of in-game characters to train the imitator and pre-train the translator. Fig.~\ref{fig:training_samples} shows some randomly selected samples.

\end{document}